\documentclass[letterpaper, 10 pt, conference]{ieeeconf}  

\IEEEoverridecommandlockouts                              

\usepackage{graphicx}   
\usepackage{svg}        
\usepackage{amsmath}    
\usepackage{amssymb}    
\usepackage{xcolor}
\usepackage{cite}
\usepackage{makecell}
\usepackage{balance}

\title{\LARGE \bf
AutodidactWAM: Cross-Modal Self-Distillation  \\ from Generated Video to Robot Actions
}

\author{Sergei Kurchev$^{1,2}$, Iaroslav Kolomiets$^{1,2}$, Miguel Altamirano Cabrera$^{1,2}$,
       Artem Lykov$^{2,1}$, and Dzmitry Tsetserukou$^{1}$
\thanks{$^{1}$Skolkovo Institute of Science and Technology (Skoltech), Bolshoy
       Boulevard 30, bld.~1, Moscow 121205, Russia.}%
\thanks{$^{2}$MWS R\&D Center.}%
\thanks{{\tt\small \{Sergei.Kurchev, Yaroslav.Kolomiyets, Miguel.Altamirano,
       Artem.Lykov, D.Tsetserukou\}@skoltech.ru}}
\thanks{Corresponding author: {\tt\small Sergei.Kurchev@skoltech.ru}}%
}

\begin{document}
\bstctlcite{IEEEexample:BSTcontrol}

\maketitle
\thispagestyle{empty}
\pagestyle{empty}

\begin{abstract}

World-action models (WAMs) such as Cosmos~3 jointly generate future video and robot actions from an observation and instruction. Adapting one such model, with a lightweight LoRA fine-tune, to a robot absent from pre-training, a Unitree~G1 humanoid with five-fingered BrainCo hands, exposes a video--action asymmetry: the video already renders plausible task executions, while the co-generated action is systematically mis-targeted. We score closed-loop real-robot trials at three cumulative stages: \emph{pre-grasp} (hand reaching a graspable pose), grasp, and pick-and-place, where the native action, on pilot objects, succeeds only $\approx\!17\%$/$10\%$/$7\%$ of the time, worse still on held-out objects. We propose \emph{AutodidactWAM}: a hand-pose estimator trained without teleoperation, followed by inverse kinematics, run on the model's own generated video to recover an action estimate $\hat a$ through those same three stages. Paired with the native prediction $a^{-}$, the resulting preferred target $a^{+}$ fine-tunes only the action-related layers, with the training loss evaluated on action outputs while the generated video is teacher-forced. We compare two ways to use this pair: plain supervised relabeling and a rectified-flow adaptation of Diffusion-DPO, rather than committing to either. After one-time embodiment adaptation, self-distillation requires no additional task-specific teleoperation. The recovered-action $\hat a$ gate reaches $\approx\!75\%$/$47\%$/$42\%$ pre-grasp/grasp/pick-and-place success, against the native $17\%$/$10\%$/$7\%$. A hybrid self-distillation objective combining preference supervision, supervised target fitting, and Cartesian trajectory anchoring (DPO+SFT+DTW) produces the strongest policy among the configurations tested: on Oreo, our training object, it reaches $90\%$ pre-grasp and $20\%$ full-task success; on the held-out object, it reaches $80\%$ and $30\%$. Plain Flow-DPO, trained without any additional losses, reaches $0\%$ success everywhere despite $1.000$ validation preference accuracy—evidence that it is the combination of terms, not the contrastive objective alone, that drives the gains above.

\end{abstract}

\section{INTRODUCTION}
\label{sec:intro}

\begin{figure}[t]
  \centering
  \includegraphics[width=0.95\linewidth]{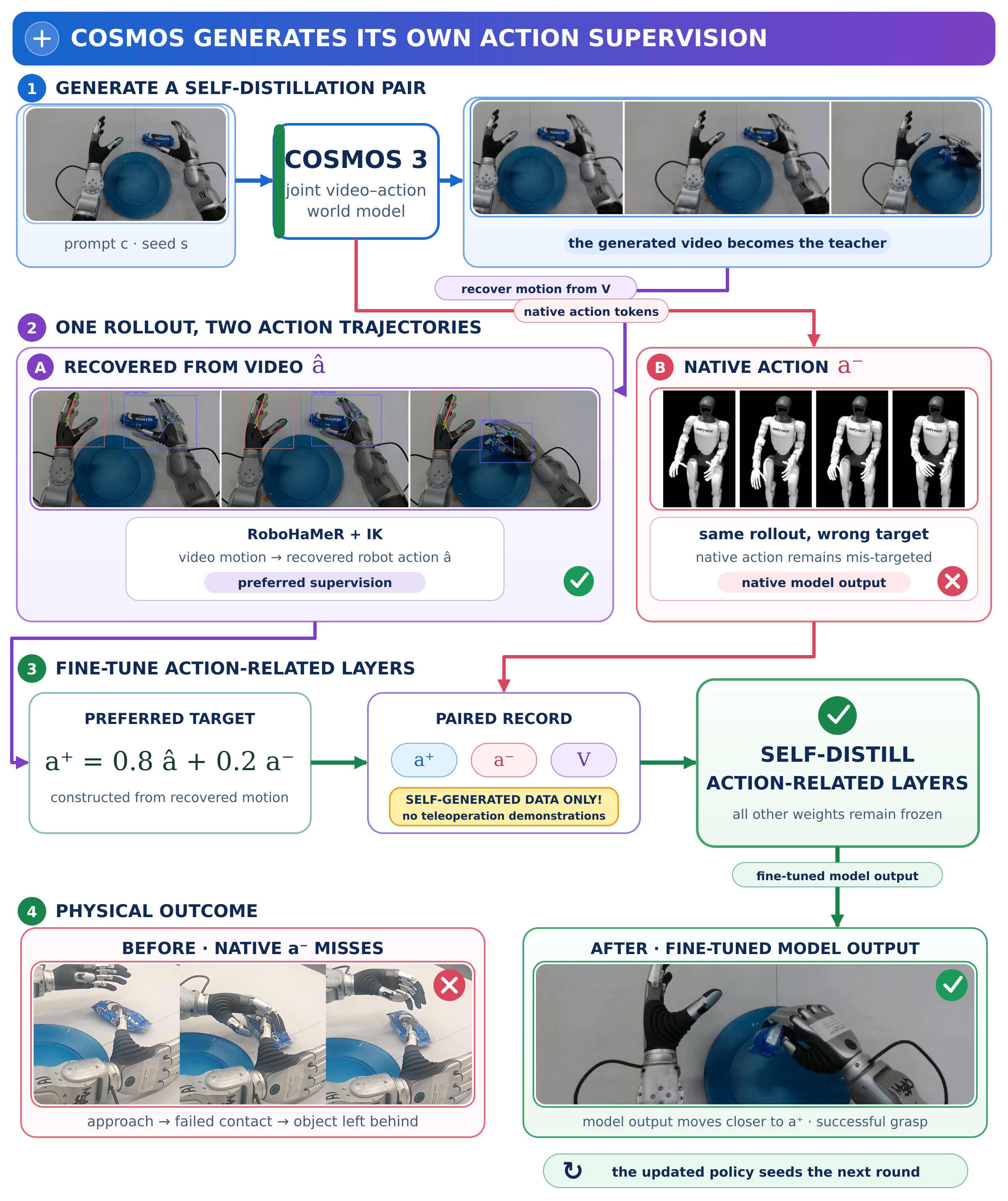}
  \caption{Cosmos generates its own action supervision. \textbf{1. generate a self-distillation pair:} one anchor frame, prompt, and seed are passed through Cosmos~3; the generated video becomes the teacher for the next step. \textbf{2. One rollout, two action trajectories:} (A) RoboHaMeR${+}$IK read the generated video itself to recover an action trajectory $\hat a$ --- the preferred supervision; (B) the same joint generation's native action tokens $a^{-}$ remain mis-targeted despite sharing the same rollout. \textbf{3. Fine-tune action-related layers:} the preferred target   $a^{+}=0.8\,\hat a + 0.2\,a_0$ and the rejected $a^{-}$, together with the teacher-forced video, self-distill only the action-related layers, with every other weight frozen and no teleoperation demonstrations involved. \textbf{4. Physical outcome:} before fine-tuning the native prediction approaches the object but fails contact and leaves it behind; after fine-tuning the same model's output moves toward $a^{+}$ and can complete the grasp in successful trials, and the updated policy seeds the next round.}
  \label{fig:teaser}
\end{figure}

Progress on generalist robot manipulation is gated by action-labeled data: teleoperation is slow, robot-specific, and must be repeated for every new embodiment~\cite{zhao2023aloha,chi2024umi,oxe2023}, hardest for dexterous multi-fingered hands. \emph{World-action models} (WAMs) address this via large-scale pre-training that predicts both future video and actions from vision and language~\cite{cosmos1,cosmos3,gr00tn1,gr1,unipi}. Cosmos~3~\cite{cosmos3} is one such model: video and action tokens are noised and denoised \emph{jointly} within one rectified-flow sampling trajectory~\cite{liu2023rectified,lipman2023flowmatching} (Sec.~\ref{sec:prelim}).

We adapt Cosmos~3 to a Unitree~G1 humanoid with five-fingered BrainCo hands, an embodiment absent from pre-training, via a lightweight LoRA fine-tune on a mixed in-house G1 corpus. The two output modalities transfer very unequally: the generated \emph{video} is usually visually plausible and temporally coherent, although occasional hallucinations remain and are characterized separately (Sec.~\ref{sec:screening}); the co-generated \emph{action} output looks plausible on its own but is systematically mis-targeted in space, completing the task in only $\approx\!7\%$ of closed-loop pick-and-place trials and reaching pre-grasp in just $\approx\!17\%$ (Table~\ref{tab:pilot}).

This suggests task-consistent information is represented more effectively in the generated video than in the co-generated action tokens, and more fine-tuning does not fix it: SFT loss decreases and DPO validation preference accuracy saturates at $1.000$ within the first round, while the on-policy task metric that tracks real behavior --- headline wrist DTW/path --- does not follow (Fig.~\ref{fig:losses}, Sec.~\ref{sec:analysis}). This points to a residual cross-modal consistency problem between the video and action outputs of one generation, related to a \emph{forward--inverse asymmetry}~\cite{wav2026}, an \emph{executability gap}~\cite{eva2026}, or a mismatch between plausible video and control-relevant dynamics~\cite{vampo2026} in concurrent work. We recover that correspondence with a frozen perception stack --- hand detection~\cite{yolo}, a hand-pose model~\cite{hamer} adapted once to the robot hand, and inverse kinematics --- run on each generated rollout to give a full action trajectory $\hat a$, paired with the native prediction $a^{-}$ to fine-tune the action-related layers toward the resulting target $a^{+}$ (Sec.~\ref{sec:impl}).

We compare two ways to use this pair (Sec.~\ref{sec:obj}): plain supervised relabeling onto $a^{+}$, and a rectified-flow adaptation of Diffusion-DPO~\cite{wallace2024diffusiondpo} that prefers $a^{+}$ over $a^{-}$, since token-level DPO~\cite{rafailov2023dpo} needs a tractable likelihood actions do not have. The generated video is teacher-forced as clean conditioning in both cases, so gradients update only the action-related layers and cannot degrade video quality; using the model's own prediction as the rejected sample mirrors self-play preference optimization~\cite{spin2024}. Every training example needs only a composited anchor frame; seeds and LLM-generated prompt variations~\cite{leo2024} turn it into an effectively unbounded stream of $(\text{video}, \hat a, a^{-})$ tuples with no manual annotation, in the lineage of self-training and self-distillation~\cite{star2022,noisystudent2020,spin2024} applied here across the outputs of one model. The perception stack itself carries only a one-time training cost from synthetic renders (Sec.~\ref{sec:extractor}), so the method adds no further teleoperation as it scales; relative to recent work on the same gap, we repair the action output offline rather than with reinforcement learning~\cite{wav2026,eva2026,vampo2026,wmpo2025} (Sec.~\ref{sec:related}).

\textbf{Contributions.}
\begin{itemize}
  \item We identify and quantify a \textbf{video--action generalization   asymmetry} in a jointly-trained WAM adapted to an unseen dexterous humanoid   embodiment: the generated video renders the task on most seeds, while the   co-generated action output produces plausible-looking but mis-targeted   motion, reaching the pre-grasp pose only $\approx\!17\%$ of the time and completing   the full pick-and-place task in $\approx\!7\%$ of closed-loop trials on the real   robot (Table~\ref{tab:pilot}).
  \item We propose \textbf{AutodidactWAM}, which recovers executable   action targets from the model's own generated video and uses them to align the   action prediction for the adapted embodiment, without collecting additional   task-specific teleoperation (Sec.~\ref{sec:obj}, \ref{sec:teleopscope}).
  \item We introduce a \textbf{scalable synthetic data engine} that generates   self-distillation pairs from a small set of static observations through seed,   prompt, and object-placement diversification.
\end{itemize}

\section{RELATED WORK}
\label{sec:related}

\textbf{VLA policies and video-as-control.}
Action policies trained on heterogeneous robot corpora~\cite{oxe2023,rt2,openvla,gr00tn1}, often with a diffusion action head~\cite{chi2023diffusionpolicy}, still generally need demonstrations \emph{on} a robot outside the training mixture; our starting point --- Cosmos~3 with a lightweight LoRA on the new embodiment --- sits in this family, and our contribution begins where it usually ends, when the adapted action branch still does not produce competent behavior. Separately, world models have served as planners or manipulation pretraining, with a separate policy or inverse-dynamics model turning predicted frames into actions~\cite{unipi,avdc,gr1,genie,vpt2022,lapa2024,doasido2025}. We differ in that video and action are two outputs of \emph{one} model and the generated video repairs that same model's action branch rather than driving a downstream controller; the recovered target $a^{+}$ is used contrastively against the model's own $a^{-}$, not as a plain regression target.

\textbf{The video--action gap, and preference optimization.}
Several recent methods target the same mismatch: WAV~\cite{wav2026} frames it as a \emph{forward--inverse asymmetry} with a subgoal generator and a sparse inverse model; EVA~\cite{eva2026} names it the \emph{executability gap} and RL-tunes the video branch against an inverse-dynamics reward trained on real robot data; VAMPO~\cite{vampo2026} and WMPO~\cite{wmpo2025} apply policy optimization to sharpen visual dynamics or run in-world-model RL for a VLA policy. We differ on four points: we correct the \emph{action} branch and leave video untouched; the optimization is offline and preference-based, with no reward model, no training-time rollouts, and no real-robot inverse-dynamics data; the extractor is a hand-pose reconstructor with IK, for the dexterous contact-rich regime where low-dimensional inverse models are weakest; and every signal is a contrastive pair against the model's own $a^{-}$ for the same generation. That contrastive step builds on DPO~\cite{rafailov2023dpo} and its diffusion extension~\cite{wallace2024diffusiondpo} (Sec.~\ref{sec:obj} adapts the latter to rectified flow) and on self-improvement from a model's own outputs~\cite{noisystudent2020,star2022,spin2024}; AutodidactWAM applies this \emph{across modalities} of one model, the stronger (video) supervising the weaker (actions).

\section{METHOD}
\label{sec:method}

\begin{figure*}[t]
  \centering
  \includegraphics[width=0.95\textwidth]{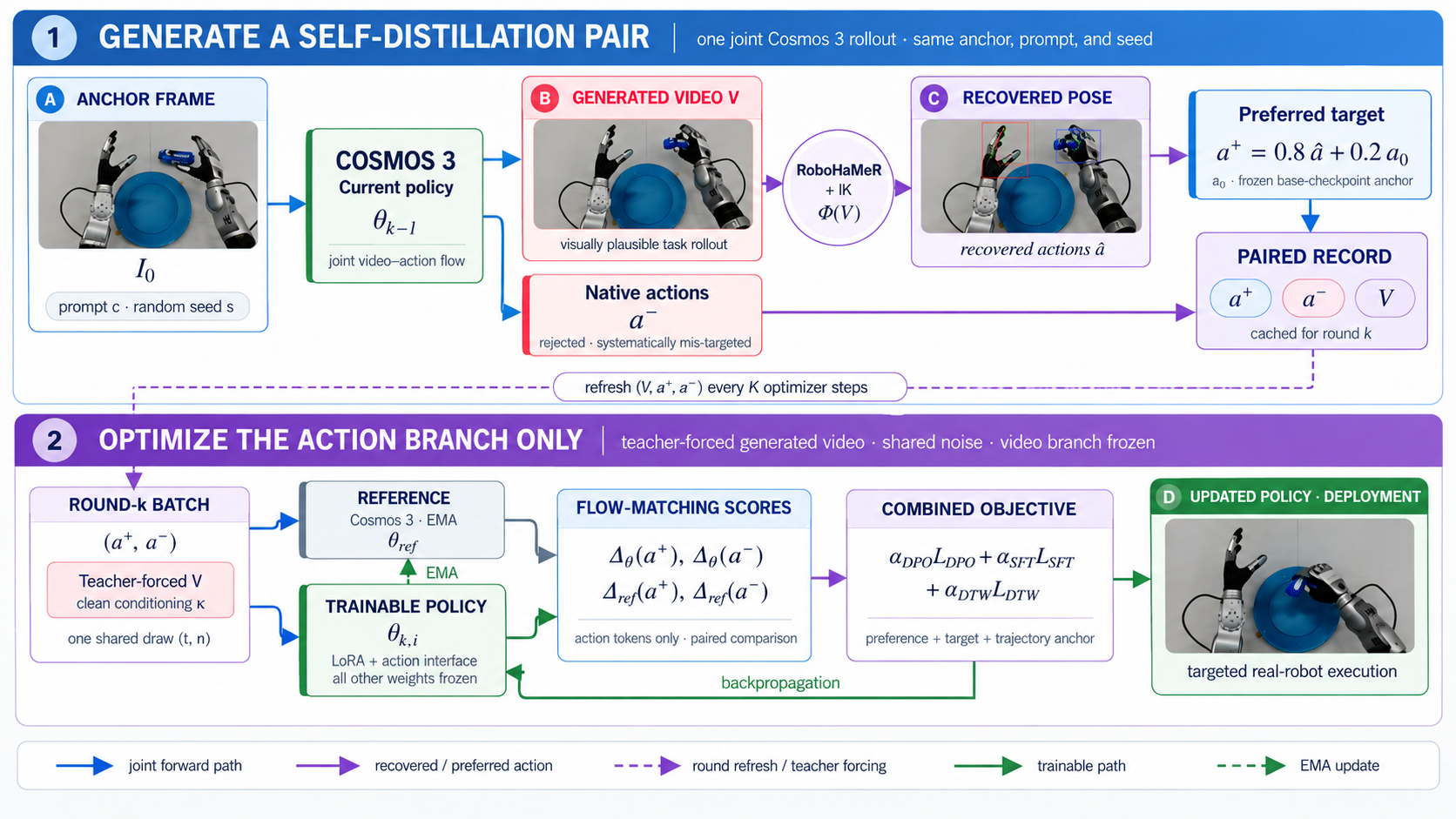}
  \caption{The AutodidactWAM pipeline. \textbf{1, generate a   self-distillation pair} (one joint Cosmos~3 rollout): from anchor frame   $I_0$ and prompt/seed, the round-$k$ policy $\theta_{k-1}$ (A) jointly   produces video $V$ (B) and native action $a^{-}$; $V$ is passed through   RoboHaMeR${+}$IK, $\Phi(V)$, to recover the pose (C) and the trajectory   $\hat a$, mixed with the round-starting prediction $a_0$ into the target   $a^{+}=0.8\,\hat a + 0.2\,a_0$; the pair is refreshed from the updated   policy every $K$ steps (Sec.~\ref{sec:pair}). \textbf{2, optimize the   action-related layers only} (teacher-forced video, shared noise): the EMA   reference $\theta_{\mathrm{ref}}$ and the trainable policy $\theta_{k,i}$   (LoRA${+}$action interface) score $(a^{+},a^{-})$, giving flow-matching   residuals $\Delta_\theta,\Delta_{\mathrm{ref}}$; the combined objective   $\alpha_{\mathrm{DPO}}\mathcal{L}_{\mathrm{DPO}} +   \alpha_{\mathrm{SFT}}\mathcal{L}_{\mathrm{SFT}} +   \alpha_{\mathrm{DTW}}\mathcal{L}_{\mathrm{DTW}}$ (Eq.~\eqref{eq:anchor}) is   backpropagated into the policy, which EMA-updates the reference; the   updated policy (D) is deployed for real-robot execution.}
  \label{fig:pipeline}
\end{figure*}

AutodidactWAM has three stages (Fig.~\ref{fig:pipeline}). First, from one anchor frame and a language instruction the world model generates a video rollout together with its native action prediction. Second, a frozen perception stack reads the \emph{generated} video and recovers a second, independent action trajectory. Third, the model's action branch is fine-tuned to prefer the recovered trajectory over its own native one, with a preference objective adapted to the model's rectified-flow formulation and with the generated video teacher-forced so that only the action tokens are updated. These stages are then \emph{iterated}: because the pair describes the action branch as it currently behaves, it is regenerated from the updated model at a fixed step interval (Sec.~\ref{sec:pair}). The remainder of this section makes each stage precise: Sec.~\ref{sec:prelim} recaps the model, Sec.~\ref{sec:pair} constructs the contrastive pair and its on-policy refresh, Sec.~\ref{sec:extractor} details the extractor that recovers $\hat a$, Sec.~\ref{sec:obj} gives the objectives, Sec.~\ref{sec:tf} explains how the generated video is teacher-forced, Sec.~\ref{sec:vidsrc} states where the generated video comes from, and Sec.~\ref{sec:impl} lists implementation details.

\subsection{Preliminaries: Cosmos~3 as a joint video--action flow model}
\label{sec:prelim}

Cosmos~3~\cite{cosmos3} is trained as a rectified flow~\cite{liu2023rectified,lipman2023flowmatching}: for clean latent $z$, noise $n\sim\mathcal{N}(0,I)$, $t\in[0,1]$, the interpolant $z_t=(1-t)z+tn$ and the network $v_\theta$ regresses the constant velocity $n-z$. \emph{One} network denoises video and action latent tokens jointly under full attention; they are not two separate branches or subnetworks, but two token types produced by the same generator, distinguished only by their own encoder/decoder and modality embedding (Sec.~\ref{sec:impl} lists exactly which action-specific tensors, plus a LoRA adapter, we fine-tune --- every layer shared with video stays frozen). For brevity, the rest of the paper still calls this action-specific slice the ``action branch,'' but it names a set of I/O tensors within one generator, not a separate network. At inference in policy mode, the model is conditioned on an anchor frame $I_0$ and prompt $c$ (no timestep embedding on conditioning tokens) and denoises video and action tokens \emph{together} along one sampler trajectory, emitting a rollout $V$ and an action chunk $a\in\mathbb{R}^{T\times57}$, $T\!\approx\!32$, for the G1's $57$-D action space (per-arm wrist pose, fingertip targets, a fixed head term), mapped by a fixed IK step to the robot's $26$ joint commands.

\subsection{From one generation to a contrastive pair $(a^{+}, a^{-})$}
\label{sec:pair}

\paragraph{The pair: $a^{-}$, $\hat a$, $a^{+}$}
A single forward pass on $(I_0, c, \text{seed})$ yields the generated video $V$ and the native action chunk $a^{-}$, denoised jointly with it. A frozen perception stack $\Phi$ (``RoboHaMeR,'' Sec.~\ref{sec:extractor}, trained without teleoperation) reads $V$ and maps the recovered joint trajectory into the same $57$-D space, giving the estimate $\hat a=\Phi(V)$ --- a reconstruction, not ground truth. The preferred target $a^{+}$ is built from $\hat a$ by a per-row choice (Sec.~\ref{sec:impl}): $a^{+}=\hat a$ for SFT, a mix-$\alpha$ interpolation with the round-starting checkpoint's own prediction for Flow-DPO. The pair is meaningful because $V$ is visually verifiable and, empirically, correct (optionally screened~\cite{robogaze2026}) while $a^{-}$ is plausible-looking but mis-targeted, so preferring $a^{+}$ needs no human annotation, mirroring self-play preference optimization~\cite{spin2024}. Because $a^{+}$ is built from an estimate, we gate on $\hat a$ directly in closed loop before any distillation (Sec.~\ref{sec:gate}; Sec.~\ref{sec:limitations} discusses residual extractor bias); $\hat a$ is resampled onto $a^{-}$'s time grid, both in normalized action units.

\paragraph{On-policy refresh}
The pair is not fixed for the whole run: distillation proceeds in rounds of $K$ optimizer steps, and at the start of each round both members are regenerated from the current checkpoint, keeping $a^{-}$ on-policy and $a^{+}$ tied to video the current model actually produces. The reference model $\theta_{\mathrm{ref}}$ (Sec.~\ref{sec:obj}) is, for all reported results, an EMA of the policy (decay $0.999$, persisted across rounds); a fully frozen reference is an untested alternative.

\subsection{Recovering $\hat a$: the RoboHaMeR extractor}
\label{sec:extractor}

$\Phi$ turns a generated RGB rollout into a joint-space trajectory through three frozen, non-contributed components, described only enough to characterize the noise floor they impose on $\hat a$ (Sec.~\ref{sec:gate}): \emph{(i)} a YOLO hand detector~\cite{yolo,yoloultralytics} (Grounding-DINO~\cite{groundingdino} fallback) gives one box per hand per frame; \emph{(ii)} RoboHaMeR, our adaptation of HaMeR~\cite{hamer} (its ViT-H/16~\cite{vitpose} encoder kept unchanged, with new heads replacing MANO's~\cite{mano}), regresses six actuator strokes, wrist rotation (6-D~\cite{zhou2019rot6d}), and metric palm depth, trained \emph{without teleoperation} on ${\sim}30$k synthetic renders of the robot swept through diverse arm/hand poses with exact, render-derived joint labels; and \emph{(iii)} a differential arm IK step~\cite{mink} drives the wrist onto the predicted pose, fingers seeded from the strokes, gaps interpolated, and the trajectory EMA-filtered into the $57$-D action space. On a held-out physical scene the actuator-stroke MAE is ${\approx}0.9\%$ and the 3-D keypoint MAE ${\approx}0.7$\,mm, bounding $\Phi$'s precision rather than its cross-scene generalization (Sec.~\ref{sec:limitations}).

\subsection{Distillation objectives}
\label{sec:obj}

Write $\Delta_\phi(a) = \big\lVert v_\phi(a_t,\,t,\,\kappa) - (n-a)\big\rVert^2$ for the rectified-flow residual of an action chunk $a$ under weights $\phi$, evaluated \emph{over the action tokens only}, with $a_t=(1-t)\,a+t\,n$ and $\kappa$ the conditioning context of Sec.~\ref{sec:tf}. Expectations below are over generations and over $(t,n)$.

\subsubsection{Supervised relabeling (SFT baseline)}
The simplest fix reuses the pretraining objective with target $a^{+}$, ignoring $a^{-}$: $\mathcal{L}_{\mathrm{SFT}}(\theta) = \mathbb{E}[\Delta_\theta(a^{+})]$. This is a mandatory control, separating ``better targets'' from ``contrastive signal'' in the final numbers.

\subsubsection{Flow-DPO}
Token-level DPO~\cite{rafailov2023dpo} does not apply --- actions have no tractable sequence likelihood --- so we adapt Diffusion-DPO~\cite{wallace2024diffusiondpo} to rectified flow, replacing its $\epsilon$-prediction error with the flow-matching residual $\Delta$. The reference $\theta_{\mathrm{ref}}$ is an EMA of the policy (decay $0.999$), starting from the same post-adaptation checkpoint as $\theta$ so the objective only re-weights toward $a^{+}$ and away from $a^{-}$. With margin $\delta_\theta(a) = \Delta_{\theta_{\mathrm{ref}}}(a) - \Delta_\theta(a)$, the Bradley--Terry objective is:
\begin{equation}
  \mathcal{L}_{\mathrm{DPO}}(\theta)
  = -\,\mathbb{E}\Big[\log\sigma\big(
      \beta\,[\,\delta_\theta(a^{+}) - \delta_\theta(a^{-})\,]
    \big)\Big],
  \label{eq:flowdpo}
\end{equation}
with $\sigma$ the logistic function and $\beta$ the DPO temperature.

Unlike SFT, which copies $\Phi$'s bias with no signal on \emph{how} $a^{-}$ is wrong, Flow-DPO scores $a^{+}$ and $a^{-}$ under identical conditioning and noise and moves only along the direction separating them, so a bias shared by both cancels in the margin. Whether this yields a real gain is the C-vs-B comparison of Sec.~\ref{sec:grid}.

\subsubsection{Cartesian target anchor (the DTW term)}
Since Eq.~\eqref{eq:flowdpo} can raise the margin by moving \emph{away} from $a^{-}$ rather than \emph{toward} $a^{+}$, we add an optional anchor tying the update to the physical target position, via the DTW alignment of predicted and target wrist trajectories (the two can drift out of phase):
\begin{equation}
  \mathcal{L}_{\mathrm{total}}(\theta)
  = \mathcal{L}_{\mathrm{DPO}}(\theta)
  + \lambda\,\mathbb{E}\big[\lVert
      \mathrm{FK}(\hat a_\theta(\kappa)) - \mathrm{FK}(a^{+})
    \rVert^2_{\mathrm{DTW}}\big],
  \label{eq:anchor}
\end{equation}
with $\mathrm{FK}$ mapping an action chunk to the absolute task-wrist position trajectory and $\lambda=0.015$.

\subsection{Teacher-forcing the video: updating the action branch only}
\label{sec:tf}

Video and action tokens are denoised in the same pass, so a naive preference step that re-noises and re-denoises the video would route preference gradients into the video branch --- exactly the component we must not disturb. We instead teacher-force the generated video: the decoded rollout $V$, re-encoded to latent tokens, enters $\kappa$ as \emph{clean} conditioning (no timestep embedding), reusing the model's existing idle-frame conditioning path, only now the entire video chunk is conditioning rather than a single frame. $\Delta_\theta$ is taken over action tokens alone; the video branch receives no gradient and the update is confined to the action encoder/decoder and the LoRA adapters. Each training step is then four action-only forward passes (policy and reference, on $a^{+}$ and $a^{-}$), far cheaper than a full joint denoising rollout.

\subsection{Where the generated video comes from}
\label{sec:vidsrc}

Every result in this submission generates $V$ with the embodiment-adapted model as it stands after its LoRA fine-tune on a mixed in-house G1 teleoperation corpus (zero-shot, ZS) --- one LoRA that trains the video and action branches jointly, after which the video branch transfers well and the action branch does not; $a^{-}$ is that model's native action trace, with no task- or scene-specific tuning added on top.

\subsection{Implementation}
\label{sec:impl}

\textbf{Models.} Base: Cosmos~3 Nano ($\sim$15.8B parameters, Qwen3-VL-8B backbone). Distillation starts from an action checkpoint already adapted to the G1 by a LoRA fine-tune over a mixed in-house teleoperation corpus, predominantly other pick-and-place tasks; its native action branch defines $a^{-}$ at the first round (Sec.~\ref{sec:experiments}). Distillation adds a rank-$64$ LoRA ($\alpha=128$) over the attention and MLP projections; the optimizer steps only the LoRA weights plus the action-interface tensors (input/output projections and the modality embedding), $16.9$M trainable parameters in all, everything else frozen. Learning rate $5\!\times\!10^{-5}$, gradient accumulation $4$, bfloat16, full activation checkpointing. This configuration is shared verbatim by every SFT and DPO variant below; what differs per variant is the round-orchestration layered on top (Sec.~\ref{sec:pair}), not the trainer itself.

\textbf{Training schedule (SFT-on-$a^{+}$, row~B).} Target is a \emph{fixed} static manifest of $105$ single-source episodes ($105\!\times\!0.8\!\approx\!84$ train / $21$ val), RoboHaMeR-extracted once and never regenerated during training, so this row carries no on-policy refresh and no reference model, and $a^{+}=\hat a$ directly with no mix-$\alpha$ blending: the loss is plain flow-matching regression toward $a^{+}$, $\mathcal{L}_{\mathrm{SFT}}(\theta)=\mathbb{E}[\Delta_\theta(a^{+})]$, no $\beta$. $500$ optimizer iterations total, checkpointed and evaluated every $100$ iterations against $20$ freshly generated on-policy episodes (generation only, no gradient) at each checkpoint, selecting on the same DTW-path metric (Sakoe--Chiba radius $32$) used by every DPO variant, so all rows are checkpoint-selected comparably.

\textbf{Training schedule (Flow-DPO, rows~C).} Rounds of $K\!=\!25$ optimizer steps, $20$ rounds, $500$ iterations total --- the same total optimizer-step budget as the SFT row above --- each generating $1$ fresh on-policy episode plus $10$ held-out validation episodes and reading every generation with RoboHaMeR at $4\times$ upscaling. $a^{+}=0.2\,a_0+0.8\,\hat a$, where $a_0$ is the frozen base checkpoint's prediction in round~$1$ only and that round's own evolving checkpoint thereafter (round~$1$'s pairs persist via anchor-pool replay); $a^{-}$ is the current checkpoint's fresh rollout. $\beta\!=\!2.0$; the reference is an EMA of the policy (decay $0.999$, persisted across rounds) for all reported results, a frozen reference being an untested alternative. Anchor-pool replay and split-clip gradient clipping (norm $1.0$, DPO and Cartesian-anchor terms independently) are active; rollback-on-regression was disabled for the reported run. Checkpoint selection uses the same DTW-path metric; both C$_{\text{zs}}$ and the combined row are trained to completion under this same $500$-iteration schedule (Sec.~\ref{sec:results}).

\textbf{Generation.} Structured action prompt in the Cosmos~3 format~\cite{cosmos3}; a composited anchor frame (Sec.~\ref{sec:dataengine}); $129$ frames, $10$ sampling steps, guidance $4.5$. Seeds and LLM-authored prompt paraphrases diversify generations (Sec.~\ref{sec:dataengine}).

\textbf{Perception.} $\Phi$ (Sec.~\ref{sec:extractor}) is applied unchanged and frozen throughout distillation; its reading of each generation, resampled onto $a^{-}$'s time grid, is $\hat a=\Phi(V)$.

\section{SYNTHETIC DATA ENGINE}
\label{sec:dataengine}

\label{sec:triple}The objectives of Sec.~\ref{sec:obj} consume tuples $(I_0, c, \mathrm{seed}) \rightarrow (V, a^{-}, \hat a)$, none requiring teleoperation or annotation. An anchor frame is built by compositing, not filming: about ten staged shots of the G1's hands over an empty table and ten photographs of the Oreo package are combined freely, the object placed at arbitrary reachable positions in software, so a dozen real captures yield an unbounded pool of anchors. Each anchor, with a structured action prompt~\cite{cosmos3}, is sent to the model in policy mode: one forward pass returns $V$ and the native $a^{-}$ (Sec.~\ref{sec:pair}); $\Phi$ reads $V$ and returns $\hat a$, cached in a manifest the pair dataset is read directly from.

\label{sec:unbounded}This supply is unbounded because three axes multiply: \emph{seeds} (reproducible per (frame, prompt)), \emph{prompts} (an LLM paraphrases the templated action prompt while keeping it valid~\cite{leo2024,actionagent2025}), and \emph{placements} (each hand pose pairs with the object anywhere reachable). The marginal cost of one more pair is one model pass plus one $\Phi$ pass --- no robot time, no human labeling; a scaling sweep is future work (Sec.~\ref{sec:limitations}).

\label{sec:teleopscope}\label{sec:screening}The starting policy is already adapted to the G1 using a mixed teleoperation corpus (Sec.~\ref{sec:experiments}); self-distillation does not remove this initial requirement, but generates all further task-specific supervision without new teleoperated trajectories. Not every generation is usable (hallucinated hand, dropped object, frozen gripper); two cheap safeguards catch this before training: anchor-pose divergence in the opening frames, and an optional vision--language plausibility judge~\cite{robogaze2026}, whose value we test by correlating its score with downstream $\hat a$ error.

\section{EXPERIMENTAL SETUP}
\label{sec:experiments}

\subsection{Task, embodiment, and checkpoints}

All experiments use a single tabletop pick-and-place task --- moving an Oreo package onto a tray --- on a Unitree~G1 humanoid with five-fingered BrainCo hands, right arm/hand only. The base model is Cosmos~3 Nano.

The \emph{start checkpoint} for all distillation runs, and row~A, is this model adapted to the embodiment by a LoRA fine-tune ($\approx\!90\%$ Oreo pick-and-place, $\approx\!10\%$ public datasets with other objects). Its native action output defines the first-round $a^{-}$, regenerated from the current checkpoint each later round (Sec.~\ref{sec:pair}); that it already completes only $\approx\!17\%$ of pre-grasp attempts and $0\%$ full pick-and-place on Oreo (Table~\ref{tab:pilot}) is part of what motivates the method. The perception stack itself uses no teleoperation (Sec.~\ref{sec:extractor}).

\subsection{Metrics and protocol}
\label{sec:metrics}

The primary metric is \textbf{closed-loop task success rate on the real robot}, since offline proxies have repeatedly diverged from real behavior on this project. We score every rollout at three cumulative stages: \emph{pre-grasp}, \emph{grasp} (implies pre-grasp), and \emph{pick-and-place} (implies both) --- separating targeting, contact, and placement failures a single success bit would conflate.

\begin{figure}[t]
  \centering
  \includegraphics[width=\linewidth]{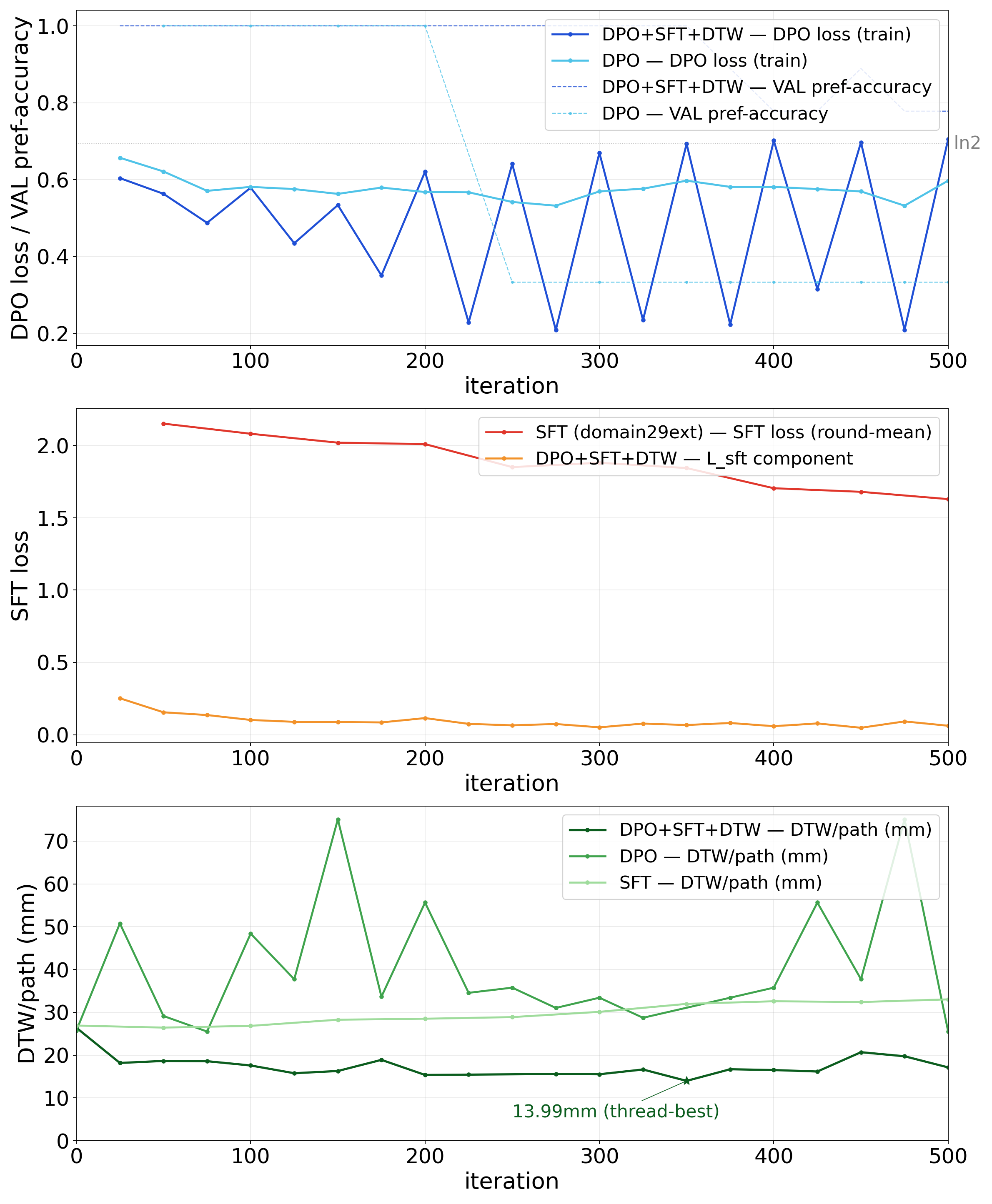}
  \caption{Training-time signal vs.\ the on-policy task metric, for the three completed runs of Sec.~\ref{sec:results}--\ref{sec:analysis} (shared $x$-axis, iterations $0$--$500$). \textbf{Top:} train DPO loss (solid) and validation preference accuracy (dashed) for DPO+SFT+DTW and plain DPO; both saturate accuracy at $1.000$ for many rounds before breaking down. \textbf{Middle:} SFT loss for plain-SFT and DPO+SFT+DTW's auxiliary term, shown only to confirm both decrease while the headline metric below diverges. \textbf{Bottom:} headline task-wrist DTW/path (mm) --- DPO+SFT+DTW is the only run separating from its round-1 base by more than noise ($13.99$\,mm thread-best at iteration $350$); plain DPO ranges $25.5$--$75.1$\,mm with no trend despite its saturated accuracy above; plain SFT improves once then degrades monotonically. Direct evidence for the training-signal/task-metric divergence of Sec.~\ref{sec:analysis}.}
  \label{fig:losses}
\end{figure}

Two offline proxies fill the gap between closed-loop evaluations: a \textbf{windowed RMSE} against the demonstrated trajectory, and the RMS gap between the \emph{absolute} wrist-position trajectories of $a^{-}$ and $a^{+}$, which catches end-effector drift a per-frame residual hides. Fig.~\ref{fig:losses} tracks this headline wrist DTW/path over training (Sec.~\ref{sec:analysis}).

\label{sec:asym}\label{sec:cost}Every row is evaluated on the \emph{same} held-out set, disjoint from training data, over $N\!=\!20$ real-robot trials per (object, condition), replayed as $\approx\!4$ consecutive $T\!\approx\!32$-step chunks executed \emph{open-loop to completion} between fresh camera captures ($\approx\!104$--$137$\,s regeneration latency per chunk). The video--action asymmetry reuses this protocol's closed-loop success for the action side (Sec.~\ref{sec:gate}); a matching video-side score has not been run (Sec.~\ref{sec:limitations}), so Fig.~\ref{fig:teaser}'s asymmetry is argued via Table~\ref{tab:pilot}. Data-engine generation averages $\approx\!8.8$\,s/episode for sampling and $\approx\!39$\,s/episode for extraction.

\subsection{Validating the recovered actions ($\hat a$ gate)}
\label{sec:gate}

Distillation is only sensible if $\hat a$ is actually better than $a^{-}$, and since self-distillation never has a matched ground-truth trajectory to compare against, we validate this directly in closed loop rather than by an offline error decomposition. The gate check replays $\hat a$ open-loop on the real robot (Video$\to$Extractor$\to$IK, no distillation) under the same three-stage protocol (Sec.~\ref{sec:metrics}); $\Phi$'s own precision on a held-out physical scene with real ground truth (Sec.~\ref{sec:extractor}) separately bounds its noise floor. In the pilot, actions recovered from zero-shot generations reach ${\approx}75\%$ real-robot pre-grasp success against ${\approx}17\%$ for $a^{-}$ (Table~\ref{tab:pilot}), clearing the gate.

\subsection{Ablation grid}
\label{sec:grid}

This submission's real-robot pilot evaluates four conditions on one held-out (anchor, prompt) set: \textbf{A}, the native $a^{-}$ baseline; \textbf{B$_{\text{zs}}$}, SFT on $a^{+}$, isolating the target alone; \textbf{C$_{\text{zs}}$}, Flow-DPO on $(a^{+},a^{-})$, isolating contrast against relabeling; and \textbf{DPO+SFT+DTW}, Flow-DPO with an EMA reference plus the SFT term and the Cartesian/DTW anchor of Eq.~\eqref{eq:anchor} together, the strongest configuration we trained. Table~\ref{tab:pilot} reports their closed-loop numbers; further objectives and video sources (Sec.~\ref{sec:obj}--\ref{sec:vidsrc}) are not evaluated here (Sec.~\ref{sec:limitations}).

The comparison that carries this submission's analysis is C$_{\text{zs}}$ vs.\ B$_{\text{zs}}$ at a fixed (ZS) video source: it isolates the contrastive objective from supervised relabeling on the same targets (Sec.~\ref{sec:results}). A further comparison --- the anchor/linear-utility variants in isolation --- was designed but not run in time (Sec.~\ref{sec:limitations}).

\section{RESULTS}
\label{sec:results}

Table~\ref{tab:pilot} reports the closed-loop pilot on three evaluation objects --- Oreo, used for self-distillation pair generation, and two object identities held out from self-distillation, a pink antistress toy and a blue ball --- at the three cumulative stages of Sec.~\ref{sec:metrics}, for the five conditions we were able to run in the time available: \textbf{Pre-DPO} (row~A, the native $a^{-}$ baseline); \textbf{Extracted $\hat a$} (Video$\to$Extractor$\to$IK, the $\hat a$ gate of Sec.~\ref{sec:gate} --- RoboHaMeR actions replayed open-loop on zero-shot video, before any distillation); \textbf{SFT} (row~B$_{\text{zs}}$, regression onto $a^{+}$); \textbf{DPO} (row~C$_{\text{zs}}$, plain Flow-DPO with no auxiliary terms); and \textbf{DPO+SFT+DTW} (Flow-DPO with the SFT-on-$a^{+}$ regularizer and the Cartesian DTW anchor of Eq.~\eqref{eq:anchor} both active). Every row past Pre-DPO shares the same trainer and LoRA configuration (Sec.~\ref{sec:impl}); DPO and DPO+SFT+DTW differ only in whether the SFT-on-$a^{+}$ regularizer and the Cartesian/DTW anchor term are zero or active. These five conditions are the full scope of this submission's real-robot pilot; the further objectives and video sources of Sec.~\ref{sec:grid} were not run in the time available and are listed as future work (Sec.~\ref{sec:limitations}), not as pending results.

\begin{table}[t]
\caption{Closed-loop pilot: success rate over $N\!=\!20$ real-robot trials per (object, condition).}
\label{tab:pilot}
\begin{center}\scriptsize
\setlength{\tabcolsep}{3pt}
\begin{tabular}{|l|c|c|c|c|c|}
\hline
\textbf{Pre-grasp} & Pre-DPO & \makecell{Extracted\\$\hat a$} & SFT & DPO & \makecell{DPO+SFT\\+DTW} \\
\hline
Oreo            & 0.2 & 0.95 & 0.7 & 0.0 & 0.9 \\
Pink antistress & 0.1 & 0.8  & 0.5 & 0.0 & 0.8 \\
Blue ball       & 0.2 & 0.5  & 0.2 & 0.0 & 0.25 \\
\hline
\textbf{Grasp} & Pre-DPO & \makecell{Extracted\\$\hat a$} & SFT & DPO & \makecell{DPO+SFT\\+DTW} \\
\hline
Oreo            & 0.1 & 0.8 & 0.15 & 0.0 & 0.3 \\
Pink antistress & 0.1 & 0.4 & 0.25 & 0.0 & 0.3 \\
Blue ball       & 0.1 & 0.2 & 0.2  & 0.0 & 0.15 \\
\hline
\textbf{Pick-and-place} & Pre-DPO & \makecell{Extracted\\$\hat a$} & SFT & DPO & \makecell{DPO+SFT\\+DTW} \\
\hline
Oreo            & 0.0 & 0.75 & 0.1 & 0.0 & 0.2 \\
Pink antistress & 0.1 & 0.3  & 0.2 & 0.0 & 0.3 \\
Blue ball       & 0.1 & 0.2  & 0.1 & 0.0 & 0.1 \\
\hline
\end{tabular}
\end{center}
\end{table}

\textbf{Cross-object transfer.} All self-distillation pairs --- for the gate check, SFT, and DPO+SFT+DTW alike --- are Oreo-only (Sec.~\ref{sec:dataengine}); the pink antistress toy and the blue ball never appear in any training generation, only at evaluation. Yet the learned correction transfers to the held-out pink object, improving pre-grasp success from $10\%$ to $80\%$ and full pick-and-place success from $10\%$ to $30\%$ (DPO+SFT+DTW vs.\ Pre-DPO). Gains on the blue ball are smaller (Sec.~\ref{sec:analysis}), indicating the current limit of this transfer.

\textbf{The $\hat a$ gate holds.} Averaged over the three objects, Extracted $\hat a$ reaches $\approx\!75\%$ pre-grasp, $\approx\!47\%$ grasp, and $\approx\!42\%$ pick-and-place success against Pre-DPO's $\approx\!17\%$ / $10\%$ / $7\%$ --- actions recovered from the model's own zero-shot generated video are markedly better-targeted than the native action branch at every stage, on objects never involved in constructing the pair (Sec.~\ref{sec:gate}), which is the precondition for distillation being worth running at all. The story this table tells: the generated video already contains a much better executable trajectory than the native action branch, and self-distillation amortizes this teacher into the action branch itself.

\textbf{SFT recovers targeting but not completion.} SFT closes much of the pre-grasp gap on Oreo and the pink toy (reaching, respectively, $0.7$ and $0.5$ against Pre-DPO's $0.2$ and $0.1$) but degrades sharply at grasp and pick-and-place relative to the gate, consistent with Sec.~\ref{sec:obj}'s prediction that regression onto $a^{+}$ inherits the extractor's own errors and competes against the model's motion prior with no signal about \emph{how} $a^{-}$ is wrong.

\textbf{DPO+SFT+DTW is the strongest trained row so far, but does not close the gate--policy gap.} It dominates plain SFT at every stage on Oreo and the pink toy (e.g.\ Oreo pre-grasp $0.9$ vs.\ SFT's $0.7$, Oreo pick-and-place $0.2$ vs.\ SFT's $0.1$) and is roughly on par on the blue ball, where all trained rows stay low --- consistent with the wrist-drift ceiling reported for this checkpoint's training run: DTW/path improves $47\%$ over its own round-1 base but plateaus after round~14, and that plateau visibly caps closed-loop success below the gate's $\hat a$ ceiling on every object.

\textbf{Plain DPO reaches $0\%$ real-robot success --- an interesting negative result.} The plain-Flow-DPO row ($\beta\!=\!2.0$, EMA reference, no SFT or Cartesian-anchor term --- Sec.~\ref{sec:impl}), trained to completion under the same $500$-iteration schedule as every other row, reaches $0\%$ closed-loop success at every stage on every object (Table~\ref{tab:pilot}), despite its validation preference accuracy saturating at $1.000$ from round~$1$ (Sec.~\ref{sec:analysis}). This is a useful data point alongside the C$_{\text{zs}}$-vs-B$_{\text{zs}}$ comparison of Sec.~\ref{sec:grid}: on this checkpoint, plain contrastive DPO with neither auxiliary term yields a policy the preference metric favors unambiguously but that fails completely in closed loop, while DPO+SFT+DTW --- adding exactly those two terms --- reaches the non-zero numbers above.

\textbf{Scope of this submission.} The five conditions of Table~\ref{tab:pilot} are the complete real-robot pilot reported in this paper; we did not have time to additionally run the further objectives and video sources of Sec.~\ref{sec:grid}. These are listed as future work (Sec.~\ref{sec:limitations}), not as results in progress.

\section{Analysis}
\label{sec:analysis}

\textbf{An interesting side finding: plain DPO's preference accuracy is a poor proxy here.} In the completed plain-DPO run, validation preference accuracy reaches $1.000$ within the first round and holds for eight rounds, while headline task-wrist DTW/path on the same checkpoints shows no trend, ranging $25$--$75$\,mm around the unchanged base. Wrist translation is represented as per-frame deltas, so a small, sign-correlated per-step error the flow-matching margin (Eq.~\eqref{eq:flowdpo}) cannot see integrates into large \emph{absolute} drift; a checkpoint can satisfy the pairwise preference almost perfectly while still drifting in Cartesian space. This is consistent with plain DPO's $0\%$ closed-loop success (Sec.~\ref{sec:results}) and with why the Cartesian/DTW anchor term was motivated in the first place, though we do not treat this single run as conclusive on its own.

\textbf{Open questions, blocked on further training runs.} Left for future compute: the paired video-branch score of Sec.~\ref{sec:asym}; a systematic search for regimes where Flow-DPO hurts beyond the blind-margin finding above, in particular reward hacking on $\Phi$ that would only surface in closed loop; why $\beta$ did not behave as a lever once gradient-norm clipping was in place; the effect of the vision--language screen of Sec.~\ref{sec:screening}; and the scaling curves of Sec.~\ref{sec:unbounded} relating data-engine size to closed-loop success.

\section{Limitations}
\label{sec:limitations}

$\hat a$ remains an estimate, not ground truth: it inherits RoboHaMeR's own domain gap and any generation artefacts (Sec.~\ref{sec:extractor}), so every distillation target carries the extractor's residual bias forward. Because $\Phi$ also defines the target Flow-DPO optimizes toward, the policy could in principle learn to satisfy $\Phi$'s output in a way that does not generalize --- a reward-hacking mode visible only in closed loop, not in any offline proxy we log. The method's reach is bounded by how far $\Phi$ itself generalizes: trained once for this embodiment and task, whether it transfers to substantially different objects or tasks without retraining is untested. Finally, every offline proxy we use --- windowed RMSE, the $a^{+}$-vs-$a^{-}$ drift readout, teacher-forced validation loss --- can miss the same failure mode Sec.~\ref{sec:analysis} documents for the preference margin, so closed-loop evaluation remains the only reliable gate for any conclusion here.

\textbf{Future work.} Screening (Sec.~\ref{sec:screening}) is currently scored only for correlation with $\hat a$ error, not yet an active filter; wiring it in, and characterizing the discarded-generations/label-noise trade-off, is future work. Every result here uses one object (Oreo); the natural scale-up is a multi-object data engine (Sec.~\ref{sec:triple}) training one policy jointly, testing whether cross-object transfer strengthens or trades off against Oreo performance. This pilot covers only four of the design points Sec.~\ref{sec:obj}--\ref{sec:vidsrc} define; the margin surrogate, the Cartesian-anchor-only and linear-utility/EMA variants, a no-embodiment-LoRA video source, and the scaling sweep of Sec.~\ref{sec:unbounded} are designed but unrun.

\section{Conclusion}

Adapting Cosmos~3 to a five-fingered humanoid absent from pre-training exposes a video--action asymmetry: the video branch of a jointly-trained world-action model renders the task convincingly while the native action branch of that same generation reaches pre-grasp only $\approx\!17\%$ of the time and completes the full task in $\approx\!7\%$ of closed-loop trials. AutodidactWAM closes part of this gap by reading the model's own generated video back through a frozen, teleoperation-free perception stack, and using the resulting estimate to fine-tune only the action branch, with no additional task-specific teleoperation collected. On the real robot, a hybrid self-distillation objective combining preference supervision, supervised target fitting, and Cartesian trajectory anchoring (DPO+SFT+DTW) produces the strongest policy among the configurations we trained, raising pre-grasp success to $90\%$ on the object used to build every training pair and, without ever seeing it during distillation, to $80\%$ on a held-out object; full pick-and-place reaches $20\%$ and $30\%$ on the two objects, respectively, against the untrained baseline's $0\%$ and $10\%$. A plain contrastive objective with neither the supervised nor the Cartesian-anchor term, by contrast, reaches $0\%$ closed-loop success despite saturating its own validation preference metric --- evidence that it is specifically the combination of terms, not contrastive (DPO) training alone, that makes self-distillation work here, and a caution against trusting offline preference accuracy as a stand-in for closed-loop behavior in this setting.





\balance
\bibliographystyle{IEEEtran}
\bibliography{references}

\end{document}